\documentclass[letterpaper]{article}
\usepackage{aaai}
\usepackage{times}
\usepackage{helvet}
\usepackage{courier}
\usepackage{latexsym}
\input{epsf}
\newtheorem{definition}{Definition}
\newtheorem{theorem}{Theorem}
\usepackage[compact]{titlesec}

\input{epsf}

\def\0{{\bf 0}}
\def\1{{\bf 1}}

\newtheorem{example}{Example}
\newcommand{\proof}[1]{\noindent{\bf Proof:} #1} 

\newcommand\jerome[1]{}

\newcommand\jomit[1]{}

\newcommand\Omit[1]{}

\begin{document}
%


\title{Dealing with Incomplete Agents' Preferences and an Uncertain Agenda in
Group Decision Making via Sequential Majority Voting}

\author{ 
 Maria Silvia Pini$^*$, 
  Francesca Rossi$^*$,  Kristen Brent Venable$^*$ and Toby Walsh$^{**}$\\
$\mbox{}^*$ University of Padova, Italy, E-mail: \{mpini,frossi,kvenable\}@math.unipd.it\\
$\mbox{}^{**}$ NICTA and UNSW Sydney, Australia, E-mail: Toby.Walsh@nicta.com.au
 }
\maketitle

\begin{abstract}
\begin{quote}
We consider multi-agent systems where agents' preferences are
aggregated via sequential majority voting: each decision is taken by performing
a sequence of pairwise comparisons  where each comparison
is a weighted majority vote among the agents.
Incompleteness in the agents' preferences is common in many real-life settings due to privacy issues
or an ongoing elicitation process.
In addition, there may be uncertainty
about how the preferences are aggregated. 
For example, the agenda 
(a tree whose leaves are labelled with 
the decisions being compared)
may not yet be known or fixed.
We therefore study how to determine collectively
optimal decisions (also called winners) when preferences
may be incomplete, and when the agenda may be uncertain. 
We show that it is computationally easy to determine if a candidate decision always wins, or may win,
whatever the agenda. On the other hand, it is computationally hard to know whether a candidate decision
wins in at least one agenda for at least one completion of the agents' preferences.
These results hold even if the agenda must be balanced so that each candidate decision faces the
same number of majority votes.  Such results are useful for reasoning about preference elicitation. 
They help
understand the complexity of tasks such as determining if a decision can be taken collectively,
as well as knowing if the winner can be manipulated by appropriately ordering the agenda.
\end{quote}
\end{abstract}


\section{Introduction}

A general method for aggregating preferences in multi-agent
systems, in order to take a collective decision, 
is running an election among the different options
using a voting rule. Unfortunately, eliciting preferences
from agents to be able to run such an election is a difficult, 
time-consuming and costly process. Agents
may also be unwilling to reveal all their
preferences for privacy reasons. 
Fortunately, we can often
determine the outcome before all the preferences
have been revealed \cite{ConitzerSandholm02b}. 
For example, it may be that one option 
has so many
votes that it  will win whatever 
happens with the remaining votes.
We can then
stop eliciting preferences.

In addition to uncertainty about the agents'
preferences, we may have uncertainty about
how the voting rule will be applied. For instance,
in sequential majority voting (sometimes called
the ``Cup'' or ``tournament'' rule), which has
been extensively studied
in Social Choice Theory \cite{Moulin91,Laslier97}, preferences are
aggregated by a sequence of pairwise comparisons.
The order of these comparisons 
(which is often called the ``agenda'') may not be fixed
or known. Nevertheless, 
we may still be able to determine
information about the outcome. 
For example, it may be that one option 
cannot win however the voting rule is applied. 
This is useful, for example, if we want to know if
the chair can control the election 
to make his 
favored option win. 

In this paper we study the computational complexity of 
determining the possible and Condorcet winners 
in sequential majority voting when 
preferences may be incomplete and/or we may not know the agenda.
We argue that the notions of possible and Condorcet 
winners considered here are
more reasonable than the earlier notions in
\cite{lprvwijcai07} as the new notions are based
on incomplete profiles as opposed to
incomplete majority graphs which potentially throw
away some information and may suggest candidates can win when they
cannot.  The old notions in \cite{lprvwijcai07} 
are upper or lower 
approximations of the new notions. 

We show that determining if an option 
always wins, or may win, in every agenda 
is polynomial. 
On the other hand, determining if an option 
wins in at least one completion of the preferences and 
at least one agenda is NP-complete.
All these results hold even if the agenda is required to be balanced.  
Because the choice of the agenda may be under the control of the chair, 
our results can be interpreted in terms of difficulty 
of manipulation by the chair (as in, e.g., \cite{Bartholdi}).



\section{Background}
\label{background}

\paragraph{Preferences.} 

We assume that each agent's preferences are specified by
a 
total order (TO) (that is, by an asymmetric,
irreflexive and transitive order)
over a set of candidates (denoted by $\Omega$). 
The candidates represent the possible options over which agents will vote. 
However, an agent may choose to reveal only partially his total order.
More precisely, given two candidates, 
say $A, B \in \Omega$, an agent specifies exactly
one of the following: $A < B$ (meaning $A$ is worse than $B$), $A > B$,
 or $A ? B$, 
where $A ? B$ means that
the relation between $A$ and $B$ has not yet been revealed.
We assume that an agent's preferences
are transitively closed. That is, if they
declare $A > B$, and $B>C$ then they also
have $A>C$. 
\begin{example}
Given candidates $A$, $B$, and $C$, an agent may
state preferences such as $A > B$, $B > C$, and $A > C$, or 
$A>B$, $B ? C$ and $A?C$.
However, an agent cannot state preferences such as
$A > B$, $B > C$, $C > A$ as this is not transitive and
thus not a total order. 
\end{example}
%

\paragraph{Profiles.}

A {\em weighted profile}
is a sequence of total orders 
describing the preferences for $n$ agents, each of which
has a given weight. 
A weighted profile is {\em incomplete} if
one or more of the preference relations is incomplete.
For simplicity,  we assume that 
the sum of the weights of the agents is odd.
An (incomplete) {\em unweighted profile}, also called  {\em egalitarian profile}, 
is one in which each agent has weight $1$. 
Given a weighted profile $P$, 
its {\em corresponding unweighted profile} $U(P)$ is the 
profile obtained from $P$ 
by replacing every ordering, say $O$,
expressed by an agent with weight $k_i$ by 
$k_i$ agents with weight $1$ all expressing 
$O$.



\paragraph{Majority graphs.} 
%


Given an (incomplete) weighted profile $P$, 
the {\em majority graph} $M(P)$ induced by $P$ is 
the directed graph whose set of vertices is $\Omega$, and where
an edge from $A$ to $B$ (denoted by $A >_m B$) denotes
a strict weighted majority of voters who prefer $A$ to $B$. 
The assumption to have an odd sum of weights ensures that there is never a tied result. 
This simplification is not  essential. We can have an even sum of weights, but in this case we have to specify 
how we deal with tied results. Thus, for simplicity we assume that the sum of weights is odd. 
A majority graph is said to be complete if, for any two vertices, 
there is a directed edge between them. 
Notice that, if $P$ is incomplete, $M(P)$ may be incomplete as well.
Also, if $M(P)$ is incomplete, the set of all complete majority graphs 
extending $M(P)$ corresponds to a (possibly proper)
superset of the set of complete
majority graphs induced by all possible completions of $P$.

\begin{figure}[th]
\begin{center}
\setlength{\epsfxsize}{2.9in}
\epsfbox{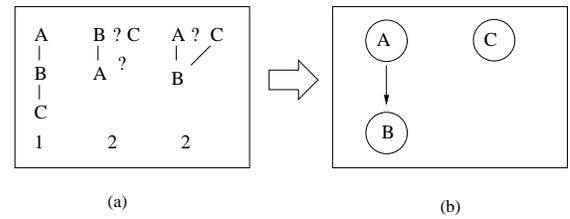}
\end{center}
\vspace*{-.3cm}
\caption{An incomplete weighted profile and its majority graph.} 
\label{mg}
\end{figure}

\begin{example}
\label{ex1}
Consider the incomplete weighted profile $P$ in Figure \ref{mg} (a). 
There are three agents $a_1$, $a_2$ and $a_3$ with weights resp. $1$, $2$, and  $2$ 
that express the following preferences: 
$a_1$ states $A > B > C$, 
$a_2$ states $B>A, A?C, B?C$ and 
$a_3$ states $A>B, A?C, C>B$. 
The majority graph induced by $P$, called $M(P)$, shown in Figure \ref{mg} (b), has 
three nodes $A$, $B$ and $C$ and one edge from $A$ to $B$, since there is 
a weighted majority of agents that prefer $A$ to $B$. There are no edges between $A$ and $C$ and between $B$ and $C$ since there are no weighted majorities that prefer one candidate to the other. 
\end{example}

\paragraph{Sequential majority voting.} 

Given a set of candidates, the sequential majority voting rule 
is defined by a binary tree (also called an {\em agenda})
with one candidate per leaf. 
Each internal node represents the candidate that wins 
the pairwise election between the node's children.
The winner of every pairwise election is
computed by the weighted majority rule, where 
$A$ beats $B$ iff there is a weighted majority of votes stating $A>B$. 
The candidate at the root of the agenda is 
the overall winner. 
Given a complete profile, 
candidates which win whatever the agenda are called {\em Condorcet winners}.

\begin{figure}[th]
\begin{center}
\setlength{\epsfxsize}{3.5in}
\epsfbox{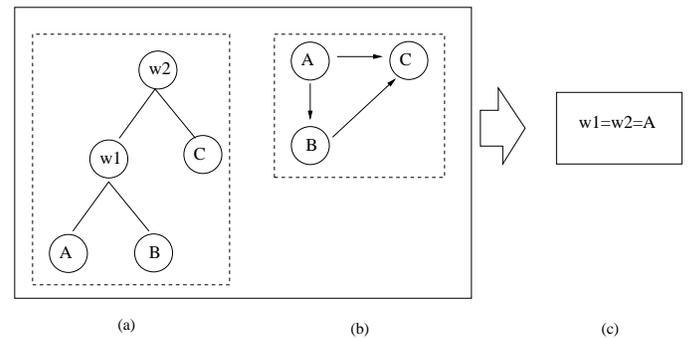}
\end{center}
\caption{How sequential majority voting works. }
\label{sq}
\end{figure}

\begin{example}
Assume to have three candidates $A$, $B$ and $C$. 
Consider the agenda  $T$ shown in Figure \ref{sq} (a). 
According to this agenda,  $A$ must first play against $B$, and then 
the winner, called $w_1$, must play against $C$. The winner, called $w_2$,  
is the overall winner. 
If we have the majority graph $M$ shown in Figure \ref{sq} (b), $w_1=A$ and $w_2=A$. 
Note that $A$ is a Condorcet winner, since it is the overall winner in all 
possible agendas. 
\end{example}

\paragraph{Winners from majority graphs.}

Four types of potential winners  
have been defined \cite{lprvwijcai07} for sequential majority voting. 
Given an incomplete 
majority graph $G$ induced by an incomplete profile $P$, consider a  
candidate $A$,  
\begin{itemize}
\item $A$ is a  {\em weak Condorcet winner\footnote{In \cite{lprvwijcai07} 
a Condorcet winner is called a necessary winner.}} for $G$ (i.e., 
$A \in WC(G)$) 
iff there is a completion of $G$ such that $A$ wins in every agenda;
\item $A$ is a {\em strong Condorcet winner} for $G$ (i.e., $A \in SC(G)$) 
iff for every completion of $G$, $A$ wins in every agenda;
\item $A$ is a {\em weak possible winner} for $G$ (i.e., $A \in WP(G)$) 
iff there exists a completion of $G$ and an agenda for which $A$ wins;
\item $A$ is a {\em strong possible winner} for $G$ (i.e., $A \in SP(G)$) 
iff for every completion of $G$
there is an agenda for which $A$ wins.
\end{itemize}


When the majority graph is complete, strong and weak Condorcet winners
coincide (that is, $SC(G) = WC(G)$). Similarly, 
strong and weak possible winners coincide in this case 
(that is, $SP(G) = WP(G)$).
In \cite{lprvwijcai07}, it is proved that 
$WP(G)$, $SP(G)$, $WC(G)$, and $SC(G)$ 
can all be computed in polynomial time.


\section{Profiles, majority graphs, and weights}

These notions of possible and Condorcet winner
are based on an incomplete majority graph. 
It is, however, often more useful and meaningful 
to start directly from the incomplete profile inducing
the majority graph. 
Given an incomplete profile, there can be more
completions of its induced majority graphs than 
majority graphs induced by completing the profile. 
An incomplete majority
graph throws away information about how
individual agents have voted. For example,
we lose information about correlations between
votes. Such correlations may prevent a 
candidate from being able to win.

\begin{example}
\label{exw}
Consider an incomplete profile $P$ with 
just one agent and three candidates ($A$, $B$, and $C$),
where the agent declare only $A>B$. 
The induced majority graph $M(P)$
has only one arc from $A$ to $B$. 
In this situation,  $B$ is a weak possible winner (that is, $B \in WP(M(P))$), since there is a 
completion of the majority graph (with arcs from $B$ to $C$ and from $C$ to $A$) 
and an agenda where $B$ wins (we first compare $A$ with $C$,
$C$ wins, and then $C$ with $B$, and $B$ wins).
However, there is no way to complete profile $P$ and
set up the agenda so $B$ wins. 
In fact, the possible completions of $P$ are $A>B>C$, $A>C>B$, and $C>A>B$, 
and in all these cases $B$ is always beaten at least by $A$. Hence, there is no agenda where 
$B$ wins. 
Note that the completion of the majority graph that allows us to conclude that $B\in WP(M(P))$    cannot be obtained in any possible completion of the agent's preferences of $P$, 
since it violates transitivity. 
Since $B$ cannot win in any completion of $P$, it is 
rather misleading to consider $B$ as a potential winner. 
\end{example}

Hence, unlike \cite{lprvwijcai07},
we will define possible 
and Condorcet winners starting directly from profiles,
rather than the induced majority graphs. 


As in \cite{ConitzerSandholm02a}, we consider
weighted votes. 
Weighted voting systems are used in 
a number of real-world settings like shareholder meetings 
and elected assemblies. 
Weights are useful in multiagent systems where we have
different types of agents. 
Weights are also interesting from a computational
perspective. 
Computing
the weak/strong possible/Condorcet winners with unweighted
votes is always polynomial. 
If there is a bounded number of candidates, 
there are only a polynomial number of
different ways to complete an incomplete profile.
Similarly, if there is a bounded number of candidates, 
there are only a polynomial number of
different ways to complete the missing links in
an incomplete majority
graph. There are also only a polynomial number
of different agendas. All the possibilities
can therefore be tested in polynomial time. 
On the other hand, adding weights to the votes
may introduce computational complexity.
For example, as we will show later, computing weak possible
winners becomes NP-hard when we add weights. 
Finally, the weighted case informs
us about the unweighted case in the presence of uncertainty about
the votes.
For instance, 
if constructive coalitional manipulation with weighted votes is
intractable,  then
it is hard to compute the probability of winning
in the unweighted case when there is uncertainty about how the votes
have been cast \cite{ConitzerSandholm02a}. 
Reasoning about weighted votes is thus closely related to 
reasoning about unweighted votes where we have
probabilities on the distribution of votes. 


\section{Possible and Condorcet winners from profiles}

We consider the following new notions of possible and Condorcet winner:

\begin{definition} 
\label{def-win}
Let $P$ be an incomplete 
weighted profile and $A$ a candidate.
\begin{itemize}
\item $A$ is a  {\em weak Condorcet winner} for $P$ (i.e., $A \in WC(P)$)
iff there is a completion of $P$ such that $A$ is a winner for all agendas;
\item $A$ is a {\em strong Condorcet winner} for $P$ (i.e., $A \in SC(P)$)
iff for every completion of $P$, and for every agenda, $A$ is a winner;
\item $A$ is a {\em weak possible winner} for $P$ (i.e., $A \in WP(P)$)
iff there exists a completion of $P$ and an agenda for which $A$ wins;
\item $A$ is a {\em strong possible winner} for $P$ (i.e.,$A \in SP(P)$)
iff for every completion of $P$ there is an agenda for which $A$ wins.
\end{itemize}
\end{definition}

It is easy to see that, 
when the profile is complete, strong and weak Condorcet winners coincide. 
The same holds also for strong and weak possible winners.

\begin{example}
Consider the profile $P$ given in Example \ref{ex1}. We have that $SC(P)$ $=$ $SP(P)$
$=\emptyset$, 
$WC(P)=\{A,C\}$, and $WP(P)=\{A,B,C\}$. 
More precisely, $A$ and $C$ are weak Condorcet winners, since there are completions of $P$
where they win in all the agendas. In fact, $A$ wins in all the agendas in the completion of
$P$ where $a_1$ states $A>B>C$, $a_2$ states $C>B>A$ and $a_3$ states $A>C>B$, while $C$ wins
in all the agendas in the completion of $P$ where  $a_1$ states $A>B>C$, $a_2$ states $C>B>A$
and $a_3$ states $C>A>B$. 
The outcome $B$ is not a weak Condorcet winner, since there are no completions where it wins 
in every agenda. However, $B$ is a weak possible winner, since there is a completion of $P$ 
and an agenda where $B$ wins
(e.g. 
$a_1$ states $A>B>C$, $a_2$ states $B>C>A$ and $a_3$ states $C>A>B$, and  $A$ first competes
 with $C$ and then the winner competes with $B$).  
Notice that in this example the weak and strong possible and Condorcet 
winners obtained considering the completions of $P$ coincide with those obtained from
considering the completions of the majority graph induced by $P$. However, as shown in Example
\ref{exw}, this is not true in general. 
\end{example}

These four notions are related to 
interesting issues in voting theory:
\begin{itemize}
\item Weak Condorcet winners are related to {\em destructive control}. 
A chair may try to build an agenda in which some candidate loses
however the votes are completed.
If a candidate is not in WC(P), then the chair can choose an agenda 
such that it must lose. Thus, the complexity of computing WC(P) 
is related to the difficulty of destructive control. 

\item Strong Condorcet winners are related to  
the {\em possibility of controlling/manipulating} the election. 
If SC(P) is non-empty, then neither the chair
nor any of the voters can do anything to change the result. 
Thus, the complexity of computing SC(P) is related to
the difficulty of manipulation/control.

\item Weak possible winners
are related to {\em participation incentives}.
If a candidate is not in WP(P), it has no chance of
winning. If it is easy for a candidate to know  
whether they are not in WP(P), he may 
drop out of the election. 
It is therefore desirable that computing WP(P) is difficult. 

\item Strong possible winners
are related to {\em constructive control}.
If a candidate is in SP(P), the chair can make the
candidate win by choosing an
appropriate agenda. Thus, it is desirable that computing SP(P) is difficult.

\end{itemize}
  
\section{Comparing the notions of winners}
\label{comp}

We now compare the notions of winners defined in 
\cite{lprvwijcai07} and those defined here.
Since in \cite{lprvwijcai07} weights were not considered, 
we first consider unweighted profiles.
%
%
%

\subsection{Unweighted profiles}
Given an incomplete unweighted profile $P$ and the incomplete 
majority graph $G$ induced by $P$, that is, $G = M(P)$,
we already observed that the completions of $G$ are a (possibly proper)
superset of the set of complete majority graphs induced 
by all possible completions of $P$. 
This observation leads to the following results.

\begin{theorem}
\label{teoG}
Given an incomplete unweighted profile $P$,
\begin{enumerate}  

\item $WP(M(P)) \supseteq  WP(P)$;

\item $SP(M(P)) \subseteq  SP(P)$;

\item $WC(M(P))= WC(P)$;

\item $SC(M(P))= SC(P)$.
\end{enumerate}
\end{theorem}

\proof{ 
Let us consider the four items separately. 
\begin{enumerate}  
\item $WP(M(P)) \supseteq  WP(P)$.  \\
If a candidate $A$ belongs to $WP(P)$, 
there is a completion of $P$, say $P'$, and an agenda, such that $A$ wins. 
Thus $A \in WP(G')$ where 
$G'$ is the complete majority graph induced by $P'$. 
Since $G'$ is one of all the possible completions of $M(P))$, 
then $A\in WP(M(P)$.  

\item $SP(M(P)) \subseteq  SP(P)$.\\ 
If a candidate is a possible winner for every completion of $G$, 
it is also a possible winner 
for the majority graphs induced by the completions of $P$, 
since they are a subset of the set of all the completions of $M(P)$. 

  
\item $WC(M(P))= WC(P)$. \\ 
Similar reasoning to the first item 
can be used to show that $WC(M(P)) \supseteq  WC(P)$. 
We can also prove that $WC(M(P)) \subseteq  WC(P)$. 
In fact, if a candidate $A$ belongs to $WC(M(P))$, 
then there must be one or more completions of
the majority graph where A has only outgoing edges.
Among such completions, there is at least one which derives from a
completion of the profile in which all $A?C$ become $A>C$ (for all $C$).
Thus, setting this is sufficient to make $A$ a weak Condorcet winner 
without contradicting transitivity of the profile.

\item $SC(M(P))= SC(P)$. \\ 
Similar reasoning to the second item 
can be used to show that $SC(M(P)) \subseteq  SC(P)$. 
We can also prove that $SC(M(P)) \supseteq  SC(P)$. 
In fact, if a candidate belongs to $SC(P)$, then it is 
a Condorcet winner, i.e., it beats every other candidate, 
for every completion of $P$. Thus it must beat 
every other candidate in the part without uncertainty. Hence,
in the (possibly incomplete) majority graph $M(P)$ induced by $P$,
there are 
outgoing edges from this candidate 
to every other candidate, and so this candidate must belong to $SC(M(P))$. 
\hfill $\Box$ 
\end{enumerate}
}

Notice that there are cases in which the subset relation
$WP(M(P)) \supseteq  WP(P)$ is strict.
In fact, a candidate can be a possible winner for a completion of $M(P)$
which is not induced by any completion of $P$,
as shown previously in Example \ref{exw}.

\subsection{Weighted profiles}

We next consider weighted profiles.
Although weighted profiles were not considered in 
\cite{lprvwijcai07}, the same notions defined there can be 
given for majority graphs induced by weighted profiles.
The analogous results to Theorem \ref{teoG} 
hold in this more general setting.
To prove this, we first show that,
given an incomplete weighted profile $P$ and  
its corresponding unweighted profile $U(P)$,  
$SC(P)=SC(U(P))$ (resp., $WC(P)=WC(U(P))$).
That is,  the set of strong (resp., weak) Condorcet winners for $P$ coincides 
with the set of strong (resp., weak) Condorcet winners for the 
unweighted profile corresponding to $P$. 
We also show that $M(P)=M(U(P))$. That is, the majority graphs of $P$ and $U(P)$ coincide.

\begin{theorem}
\label{wsc-inc}
Given 
an incomplete weighted profile $P$, 
\begin{enumerate}
\item $M(P) = M(U(P))$;
\item $SC(P)=SC(U(P))$;
\item $WC(P)=WC(U(P))$.
\end{enumerate}
\end{theorem}

\proof{ 
\begin{enumerate}
\item $M(P)=M(U(P))$.\\
The statement can be easily proven
since $U(P)$ is a profile obtained from $P$ 
by replacing each agent with weight $k_i$ and with preference ordering $O$ by 
$k_i$ agents with weight $1$ all with preference ordering $O$. 
%

\item $SC(P) = SC(U(P))$.\\
($\supseteq$)  This follows from the completions of $U(P)$ being a superset 
of the completions of $P$.\\
($\subseteq$) Assume that $A \not \in SC(U(P))$. Then $A$ does not have $m-1$ outgoing edges (where $m=|\Omega|$) in $M(U(P))$ \cite{lprvwijcai07}. Hence, since $M(P)=M(U(P))$,  $A$ does not have $m-1$ outgoing edges in $M(P)$. Hence, there is a candidate $B$ s.t. $B>_mA$ or $B?_mA$ in $M(P)$. If $B>_mA$ in $M(P)$, then for every completion of $P$ we have $B>A$,  and thus $A$ cannot win in every agenda. If $B?_mA$ in $M(P)$, then there exists a completion of $P$ where we replace every $A?B$ with $B>A$, where $A$ may not win. Hence $A$ does not win in every completion and agenda.

\item $WC(P) = WC(U(P))$.\\
($\subseteq$) This follows from the completions of $U(P)$ being a superset of the completions of $P$.\\
($\supseteq$)  Assume that $A \in WC(U(P))$.Then $A$ has no ingoing edges in $M(U(P))$ \cite{lprvwijcai07}. Hence, since $M(U(P))=M(P)$, $A$ has no ingoing edges in $M(U(P))$. Thus, if we replace, for every $B$, $A?B$ in $P$ with $A>B$, we obtain a completion of $P$ where $A$ wins in every agenda. Hence $A\in WC(P)$. \hfill $\Box$ \\
\end{enumerate} 
}
%

We can now compare the notions of winners in the weighted case.

\begin{theorem}
\label{teoG1}
Given an incomplete weighted profile $P$, 
\begin{enumerate} 

\item $WP(M(P)) \supseteq WP(P)$, 
that is, the set of the weak possible winners for the majority graph induced by $P$ contains or is equal to the set of the weak possible winners for $P$;

\item $SP(M(P)) \subseteq SP(P)$,
that is, the set of the strong possible winners for the majority graph induced by $P$ is contained  or is equal to the set of the strong possible winners for $P$;

\item $SC(M(P))=SC(P)$,
that is, the set of the strong Condorcet winners for the majority graph induced by $P$ is equal to the set of the strong Condorcet winners for $P$; 

\item $WC(M(P))=WC(P)$,
that is, the set of the weak Condorcet winners for the majority graph induced by $P$ is equal to the set of the weak Condorcet winners for $P$.  

\end{enumerate} 
\end{theorem}

\proof{
Let $U(P)$ be the unweighted profile obtained from $P$. 
\begin{itemize}
\item 
1st and 2nd item:\\
Since the completions of $P$ 
are a subset of the completions of $U(P)$, 
$WP(P) \subseteq WP(U(P))$ and $SP(P) \supseteq SP(U(P))$. 
Now, since $M(P)=M(U(P))$ by Theorem \ref{wsc-inc}, and since 
$SP(G)$ and $WP(G)$ depend only on the majority graph $G$ 
under consideration, 
$WP(M(P)) \supseteq WP(P)$ and $SP(M(P)) \subseteq SP(P)$. 
\item 3rd and 4th item:\\ 
To prove that $SC(P) = SC(M(P))$, we may notice that
$SC(P) = SC(U(P))$ by Theorem \ref{wsc-inc},
$SC(U(P)) = SC(M(U(P)))$ by Theorem \ref{teoG},
and $SC(M(U(P))$ $ = SC(M(P))$ by Theorem \ref{wsc-inc} and by the fact that 
$SC(G)$ depends only on the majority graph $G$ considered.
The same reasoning allows us to conclude that 
$WC(P)=WC(M(P))$. \hfill $\Box$ 
\end{itemize}
}

Note that Theorems 1 and 3 show that the same
relationships hold with or without weights.
It is perhaps interesting to observe
that a stronger relationship cannot be shown
to hold in the more specific case of unweighted
votes.

\section{Complexity of determining winners}

We now turn our attention to 
the complexity of determining possible and 
Condorcet winners from profiles.
We start by showing that computing the 
weak or strong Condorcet winners
is polynomial in the number of agents and candidates.

%

\begin{theorem}
\label{wc-pol}
Given an incomplete weighted profile $P$, 
the sets $WC(P)$ and $SC(P)$ 
are polynomial to compute.
\end{theorem}

\proof{ 
By Theorem \ref{teoG1}, 
$WC(P)$ $= WC(M(P))$ and $SC(P)$ $= SC(M(P))$.
Moreover, by Theorem \ref{wsc-inc} we know that 
$M(P)=M(U(P))$, where $U(P)$ is 
the corresponding unweighted profile obtained from $P$.  
Thus, we have that $WC(P)$ $= WC(M(U(P)))$ and $SC(P)$ $ = SC(M(U(P)))$.
In \cite{lprvwijcai07} the authors show that, 
given any majority graph $G$ obtained from an unweighted profile, 
it is polynomial to compute $WC(G)$ and $SC(G)$.
Hence,  it is  polynomial to compute $WC(M(U(P)))$ and $SC(M(U(P)))$.
\hfill$\Box$ \\
}


Since, as noted above, WC(P) is 
related to destructive control and SC(P) is related to 
the possibility of control or manipulation, this means that:
\begin{itemize}
\item It is easy for a chair to control
destructively the election. That is,
given a candidate $A$, it is easy for the chair 
to know whether, no matter how votes will be completed,
there is an agenda where $A$ does not win.
\item It is also easy for a chair or a voter to know 
whether control/manipulation is possible.
\end{itemize}


We next show that
computing weak possible winners is intractable in general. 


\begin{theorem}
\label{toby}
Given an incomplete weighted profile $P$ with 3 or more
candidates, deciding if a candidate is in WP(P) 
is NP-complete.
\end{theorem}

\proof{
Clearly the problem is in NP as a polynomial 
witness is a completion and an agenda in which the candidate
wins. To show it is NP-complete, 
we give a reduction from the number partitioning
problem. 
The reduction is based around constructing a
Condorcet cycle and is similar to those 
used in \cite{ConitzerSandholm02a} to show that manipulation
is computationally hard even with a small number of
candidates when votes are weighted.
The reduction is, however, different to that
in Theorem 8 in \cite{ConitzerSandholm02a} as the reduction 
there concerns the randomized Cup rule and 
requires 7 or more candidate.

We have a bag of integers, $k_i$ with sum $2k$ and
we wish to decide if they can be partitioned into
two bags, each with sum $k$. 
We want to show that a candidate $B$ is a weak possible winner 
if and only if such a partition exists.
We construct an incomplete profile over three candidates
($A$, $B$, and $C$) as follows. We have 
$1$ vote for $B>C>A$ of weight $1$,
1 vote $B>A>C$ of weight $2k-1$, 
and 1 vote $C>B>A$ of weight $2k-1$. 
At this point, 
the total weight of votes with $B>A$ exceeds that of $A>B$ by $4k-1$,
the total weight of votes with $B>C$ exceeds that of $C>B$ by $1$, and 
the total weight of votes with $C>A$ exceeds that of $A>C$ by $1$.

We also have, for each $k_i$, a partially specified vote of weight $2k_i$
in which we know just that $A>B$.
As the total weight of these partially
specified votes is $4k$, we are sure $A$ beats $B$ in the final
result by 1 vote. 
The only agenda in which $B$ can win
is the one in which $A$ plays $C$ and the winner
then plays $B$. In addition, 
for $B$ to win, the partially specified votes need to
be completed so that $B$ beats $C$, and $C$ beats $A$ in the final result. 
We show that this is possible iff
there is a partition of size $k$. 
Suppose there is such a partition. Then let the votes
in one bag be $A>B>C$ and the votes in the other
be $C>A>B$. Then, $A$ beats $B$,
$B$ beats $C$ and $C$ beats $A$, all by 1 vote in the final result.
On the other hand, suppose there is a way to cast
the votes to give the result
$A$ beats $B$,
$B$ beats $C$ and $C$ beats $A$. 
All the uncast votes rank $A$ above $B$. 
In addition, at least half the weight of votes must rank $B$ above
$C$, and at least half the weight of votes must rank $C$ above $A$. 
Since $A$ is above $B$, $C$ cannot be both above $A$ and below
$B$. Thus precisely half the weight of votes 
ranks $C$ above $A$ and half ranks $B$ above $C$. Hence
we have a partition of equal weight. 
We therefore conclude that $B$ can win iff there is
a partition of size $k$. 
That is, deciding if $B$ is a weak possible winner is NP-complete. 
We can extend the reduction to more than 3 candidates
by placing any additional candidate at
the bottom of every voters' preference ordering 
(it does not matter how). 
\hfill $\Box$\\
}

Note that computing weak possible winners from an incomplete
majority graph is polynomial \cite{lprvwijcai07}. 
Thus, adding weights to the votes and computing weak possible
winners from the incomplete profile instead of the majority
graph makes the problem intractable. On the other hand, adding weights to the
votes did not make weak and strong Condorcet winners harder to compute. 

We recall that $WP(P)$ is related to participation incentives.
Thus Theorem \ref{toby} tells us that it is difficult for a candidate to know 
whether they have chance to win. 
This makes it less probable that they drop out. 

Also, $WP(P) \subseteq WP(M(P))$ (see Theorem \ref{teoG1}).
Thus, while $WP(P)$
is difficult to compute, it is easy to compute a superset of it, that is, $WP(M(P))$. 


The complexity of determining strong possible winners from an incomplete profile
(that is, the set $SP(P)$) remains an open problem. However, we know that 
computing strong possible winners from incomplete majority graphs (that is,
$SP(M(P))$) without weights is easy \cite{lprvwijcai07}. 
This gives us an easy way to compute a subset of $SP(P)$:
if $SP(M(P))$ is not empty, we can easily 
compute it and find at least some of the candidates in $SP(P)$.






\section{Fair possible and Condorcet winners}

All the notions of winners defined so far 
consider agendas of any shape. 
Agendas that are unbalanced 
may not be considered "fair". Such agendas may allow 
weak candidates, that can beat only a small number 
of candidates, to end up winning the election.
We therefore consider, as in \cite{lprvwijcai07},
agendas that are balanced binary trees.
 
Given a complete profile $P$,
a candidate $A$ is said to be a {\em fair possible winner} 
for $P$ iff there is a  balanced agenda in which $A$ wins.
A balanced agenda is a binary tree in which the difference between the
maximum and the minimum depth among the leaves is less
than or equal to 1. 
%
Testing whether a candidate is a fair possible winner 
over weighted majority graphs is NP-hard
\cite{lprvwijcai07}.

\begin{figure}[th]
\begin{center}
\setlength{\epsfxsize}{3.0in}
\epsfbox{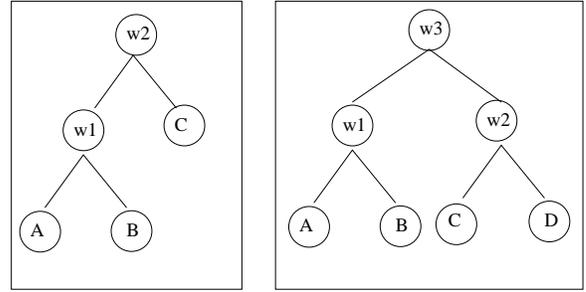}
\end{center}
\caption{Two balanced agendas.} 
\label{bal}
\end{figure}

\begin{example}
Figure \ref{bal} shows two balanced agendas. 
\end{example}

We now apply this notion of fairness to our definition of winners based
on incomplete profiles.
Thus, given an incomplete weighted profile 
$P$, we define {\em fair strong Condorcet} (FSC(P)), 
{\em fair weak Condorcet} (FWC(P)), {\em fair strong possible} (FSP(P)), 
and {\em fair weak possible} (FWP(P)) winners
in an analogous way to Definition \ref{def-win}
but limited to fair agendas.
For example, a candidate is in FSC(P) iff 
they win in all completions of profile P and in
all {\em balanced} agendas. 




We now show that it is easy to compute 
fair weak Condorcet or fair strong Condorcet winners
based on the observation that 
fairness 
does not change these sets.

\begin{theorem}
Given an incomplete weighted profile $P$, 
\begin{itemize}
\item $FSC(P)=SC(P)$ and $FWC(P)=WC(P)$;  
\item $FWC(P)$ and $FSC(P)$ are polynomial to compute.
\end{itemize}
\end{theorem}

\proof{ 
We first show that $FSC(P)=SC(P)$ (resp., $FWC(P)=SW(P)$). \\
($\supseteq$) If $A \in SC(P)$ (resp., $WC(P)$), for every completion (resp., for some completion) of $P$, $A$ wins in   
every agenda. Thus $A$ wins also in balanced agendas. 
Hence, $A \in FSC(P)$ (resp., $A \in FWC(P)$). \\
($\subseteq$) If $A \in FSC(P)$ (resp., $FWC(P)$), 
for every completion (resp., for some completion) of $P$, $A$ wins in  
every balanced agenda.
In every balanced agenda, $A$ must win against at least a candidate
(the one in $A$'s first match).
If $A$ wins in every balanced agenda,
it therefore means that $A$ must win against every candidate.
Thus $A$ wins in every agenda.
Thus, $A \in SC(P)$ (resp., $A \in WC(P)$). 

Since $FSC(P)=SC(P)$ and  $FWC(P)=SW(P)$, and since, by Theorem \ref{wc-pol},
$SC(P)$ and $WC(P)$ are polynomial to compute, also 
$FSC(P)$ and $FWC(P)$ are polynomial to compute. \hfill $\Box$\\
}

Thus, the test for destructive control (related to WC) 
and for the possibility of control/manipulation (related to SC) 
are easy even when we consider only fair agendas.

Let us now consider fair weak possible winners.
Since every balanced agenda is also an agenda, we have that 
$FWP(P) \subseteq WP(P)$.  
We already know from Theorem \ref{toby}
that determining WP(P) is difficult.
We will now show that this remains so for FWP(P).
 
\begin{theorem}
Given an incomplete weighted profile $P$ with 3 or
more candidates, 
deciding if a candidate is in FWP(P) 
is NP-complete.  
\end{theorem}

\proof{
We use the same construction
as in the proof of Theorem \ref{toby}. 
Given the profile constructed there, 
the only possible fair agendas in which 
B wins are those in which A plays 
C, and (at some later point) B then plays the winner.
All the additional candidates will be defeated by A, B and C so
can be placed anywhere in the fair agenda. \hfill $\Box$\\
}

Since the notion of weak possible winner is related to the 
concept of losers (losers are those not in WP), this means that
it is difficult to know whether a candidate is a loser (or 
alternatively still has a chance to win).
This difficulty remains so even if we consider only balanced agendas.

The computational complexity of determining fair strong possible winners
remains an open question, just as is the complexity of computing strong possible winners.
We only know that, since every balanced agenda is also an agenda,  
$FSP(P) \subseteq SP(P)$.


\section{Related work}

There has been much research on
the computational complexity of determining 
winners of various kinds for several voting rules, and of the
relationship with the complexity of problems found in preference elicitation and manipulation. 
Our results follow this same line of work while focusing on sequential majority voting.

The most related work is \cite{lprvwijcai07}
Like our paper, this
considers the computational complexity of determining 
winners for sequential majority voting. However, 
they start from an incomplete majority graph which throws away
information about individual votes, whilst we start from an incomplete profile.

Conitzer and Sandholm also consider sequential majority voting \cite{ConitzerSandholm02a}, but they assume a complete profile and a fixed agenda.  
They  show that, if the agenda is fixed and balanced, 
determining the candidates that win in at least one completion of the profile 
is polynomial, but randomizing the agenda makes deciding
the probability that a candidate wins
(and thus manipulation) NP-hard. 
They also prove that constructive manipulation  is intractable
for the Borda, Copeland, Maximin and STV rules using
weighted votes even with a small 
number of candidates. However, all of these rules
are polynomial to manipulate destructively 
except STV. 

Conitzer and Sandholm also prove that deciding
if preference elicitation is over
(that is, determining if the remaining 
votes can be cast so a given candidate
does not win) is NP-hard for the STV rule
\cite{ConitzerSandholm02b}. 
For other common voting rules like
plurality and Borda,
they show that it is polynomial
to decide if preference elicitation is over. 

The notions of possible and necessary winners are not new. 
They were introduced by 
Konczak and Lang in \cite{KonczakLang05} in the context of positional
scoring voting rules with incomplete profiles. 
A possible winner in \cite{KonczakLang05} is a candidate that can win in at least a completion of profile, while a necessary winner is a candidate that wins in every completion of the profile. We have adapted these notions to the context of sequential majority voting with complete profiles, where the unknown part is the agenda. Hence, we have defined possible winners as those candidates that may win in at least an agenda and necessary winners (called here Condorcet winners) as those candidates that win in every agenda. 
We have also considered the presence of incomplete profiles and in this case we have defined new notions of winners: weak (resp., strong) possible and necessary winners, that are those candidates that are possible and necessary winners in some (resp., in all) completions of the profile. 
We have also analyzed the complexity of determining weak and strong possible and  necessary winners from incomplete profiles for sequential majority rule, and we have shown that
 determining  weak possible winners is NP-hard, whilst determining the other kinds of winners is polynomial. 
Konczak and Lang proved that it is polynomial to compute 
both possible and necessary winners  for positional 
scoring voting rules like the Borda and plurality rule, 
as well as for a non-positional rule like Condorcet 
\cite{KonczakLang05}. 

Pini {\it et al.} 
prove that
computing the possible and necessary
winners for the STV rule is NP-hard \cite{prvwijcai07}. 
They show it is NP-hard even to approximate
these sets within some constant factor in size. 
They also give a preference elicitation procedure
which focuses just on the set of possible
winners.


Finally, Brandt {\it et al.}  consider different notions of winners starting from incomplete 
majority graphs \cite{felix}. We plan to investigate these kinds of winners in our framework.


\section{Conclusions}

We have considered multi-agent settings where 
agents' preferences may be incomplete and are aggregated 
using weighted sequential majority voting.
For this setting, we have shown that it is easy to determine
weak and strong Condorcet winners, i.e., 
to determine the candidates that win whatever the agenda, while 
it is hard to know whether a candidate is 
a weak possible winner, i.e., if the candidate
wins in at least one agenda 
for at least one completion of the agents' preferences.
This is hard even if we require that the agenda 
be a balanced tree. 
These results show that, for weighted sequential majority voting, it is 
\begin{itemize}
\item computationally easy to test if destructive control 
is possible, even if we consider only fair agendas;
\item computationally easy to test if there is a guaranteed winner, even for fair agendas;
\item computationally difficult to test if a candidate is a loser, even for fair agendas.
\end{itemize}
Our results are thus useful to understand the complexity of 
both manipulation and preference elicitation. 

The computational complexity
of testing whether constructive control is possible (that is, of
finding strong possible winners) with fair or unfair agendas remains open
and needs to be studied further.
Another interesting direction for future work
is deciding which candidates are most likely to win, which 
is related to probabilistic approaches to voting theory. 
We also plan 
to study other forms of uncertainty in the application of
the voting rule, such as uncertain weights in a scoring rule, or
a chair who can choose between different voting rules. 
We intend to analyze the presence of ties in agents' preferences. 
Adding ties requires adding a tie-breaking rule to be able to
declare a winner in each pairwise comparison.
We believe similar results can be derived for such
weak (as opposed to total) orders. 
The analysis will have to be more complex to deal with the extra cases.
However, the set of completions of the majority graph remains a 
superset of the set of completions of the profile. 
Thus all the results based on this fact still hold. 

\section{Acknowledgments}

This work has been partially supported by Italian MIUR PRIN project
``Constraints and Preferences'' (n. $2005015491$). 
The last author is funded by the
Department of Broadband, Communications and the Digital Economy, 
and the Australian Research Council. 
We would like to thank Jerome Lang for his valuable comments. 

\bibliographystyle{aaai}
\bibliography{vote-seq}

\end{document}